\documentclass[sn-basic]{sn-jnl}
\usepackage{amssymb}
\usepackage{amsmath}

\usepackage{hyperref}
\usepackage{url}
\usepackage{fancyvrb}
\usepackage{graphicx}
\usepackage{enumitem}
\usepackage[flushleft]{threeparttable}
\usepackage{multirow}

\usepackage{enumitem}
\usepackage{booktabs}
\usepackage{adjustbox}
\usepackage{rotating}

\jyear{2021}%

\theoremstyle{thmstyleone}%
\theoremstyle{thmstyletwo}%

\theoremstyle{thmstylethree}%

\begin{document}

\title[Reinforcement Based Grammar Guided Symbolic Regression]{A Reinforcement Learning Approach to Domain-Knowledge Inclusion Using Grammar Guided Symbolic Regression}


\author*[1,2]{\fnm{Laure} \sur{Crochepierre}}\email{laure.crochepierre@\{rte-france.com, univ-lorraine.fr\}}

\author[3]{\fnm{Lydia} \sur{Boudjeloud-Assala}}

\author[2]{\fnm{Vincent} \sur{Barbesant}}

\affil*[1]{\orgname{Universit\'e de Lorraine, CNRS, LORIA, F-57000}, \orgaddress{\city{Metz}, \country{France}}}

\affil[2]{\orgname{R\'eseau de Transport d'Electricit\'e (Rte) R\&D}, \orgaddress{\city{Paris}, \country{France}}}


\abstract{In recent years, symbolic regression has been of wide interest to provide an interpretable symbolic representation of potentially large data relationships. Initially circled to genetic algorithms, symbolic regression methods now include a variety of Deep Learning based alternatives. However, these methods still do not generalize well to real-world data, mainly because they hardly include domain knowledge nor consider physical relationships between variables such as known equations and units. Regarding these issues, we propose a Reinforcement-Based Grammar-Guided Symbolic Regression (RBG2-SR) method that constrains the representational space with domain-knowledge using context-free grammar as reinforcement action space. We detail a Partially-Observable Markov Decision Process (POMDP) modeling of the problem and benchmark our approach against state-of-the-art methods. We also analyze the POMDP  state definition and propose a physical equation search use case on which we compare our approach to grammar-based and non-grammar-based symbolic regression methods. The experiment results show that our method is competitive against other state-of-the-art methods on the benchmarks and offers the best error-complexity trade-off, highlighting the interest of using a grammar-based method in a real-world scenario.}




\maketitle

\section{Introduction}
Finding a generic symbolic representation from an observation set has been of a long interest in the Physics community. Long before Newton's fallen apple, scientists have been working on discovering symbolic relationships that satisfy observations, current knowledge, and already known equations. In astronomy, for example, the search for a closed-form solution has been an ongoing interest notably for Kepler's study of planetary motion and still today for the modeling of albedos \citep{Heng2021}. However, in recent years, the number of observations available has surged thanks to the increase of sensor monitoring systems in Big Data environments and, finding the best possible equation to model complex systems is now a tedious task without computer assistance. 

This automatic equation search task is known as Symbolic Regression (SR). It aims at finding a symbolic function $f$ that matches the relationship $f(X) = y$  between an observation set $X \in \mathbb{R}^n$ described with $n$ variables and a target variable $y \in \mathbb{R}$ to explain $y$ from $X$. SR was, until recently, mainly performed using Genetic Programming (GP) \citep{DBLP:conf/ppsn/Koza90}, a family of techniques that draws inspiration from Darwinian evolution to search for solutions automatically. However, the initial GP formulation is not suited for large search spaces \citep{782609}. 
To cope with this limitation, since the early works on GP \citep{DBLP:conf/ppsn/Koza90}, multiple extensions have been proposed to restrict the search space, such as Strongly Typed Genetic Programming \citep{montana1995strongly} or Grammar Guided Genetic Programming (G3P) \citep{koza2006genetic}. These methods share the interesting property of providing a way to enforce domain knowledge into the learning process \citep{DBLP:conf/ppsn/RatleS00} and are thus applied to various real-world problems \citep{DBLP:conf/cec/CherrierPDS19}.

Meanwhile, Deep Learning (DL) methods have progressively become ubiquitous because of their high representational capacity when trained on large datasets \citep{ILSVRC15}. However, in the general case, DL often has a ``black box" behavior and lack of interpretability. Thus, combining DL and SR by taking advantage of the computational capacity of DL and the expressiveness of SR would help provide more human-readable results. Along with the rapid development of DL, there have also been a growing interest in Reinforcement Learning (RL) methods to solve decision-making problems. RL is a Machine Learning paradigm inspired by behavioral psychology concerned with how an agent should act in an environment to maximize a cumulative reward \citep{sutton2018reinforcement}. Deep reinforcement learning (Deep RL) approaches especially show great power in solving complex sequential problems. Recent work in these both domains also offers viable alternatives to GP on SR tasks \citep{petersen2021deep,udrescu2020ai}. Still, unlike GP-based methods, they are not yet able to include sophisticated domain-related knowledge or constraints.

In this work, we propose a method called Reinforcement Based Grammar Guided Symbolic Regression (RBG2-SR) to tackle SR with a Deep Reinforcement Learning (Deep RL) approach using a Backus-Naur Form (BNF) \citep{DBLP:journals/cacm/Knuth64a} Context Free Grammar action space which constrains the solution space only to domain-viable solutions. Our method enforces domain-related constraints as grammatical rules in a human-readable manner. These constraints are then translated into directly interpretable symbolic outputs. We also propose a general Partially-Observable Markov Decision Process (POMDP) \citep{KAELBLING199899} modeling of the SR problem. We show that our approach performs better than other tested methods with the same grammar on the tested benchmarks. We also offer comparative results on a real-world scenario to show how our approach could use a BNF grammar to take advantage of expert knowledge.

The rest of this paper is organized as follows. First, Section \ref{sec:related_works} summarizes related state-of-the-art works. In Sections \ref{sec:proposed_approach} and \ref{sec:experiments}, we respectively describe the proposed method and its corresponding experimental results. Finally, Section \ref{sec:conclusion} offers concluding remarks and perspectives.

\section{Related Works}
\label{sec:related_works}
\subsection{Symbolic Regression}
Symbolic Regression (SR) is the process of searching for a symbolic relationship, also called \emph{expression}, that accurately matches a given dataset. Found expressions can be represented as trees where nodes contain operations and variables. 
SR was early investigated with Genetic Programming (GP) approaches \citep{DBLP:conf/ppsn/Koza90, DBLP:conf/foga/Koza92}. They iteratively evolve a population of individuals (each individual representing an \emph{expression}) across multiple generations through evolutionary operations. Many works today are descendants of these works, and propose to overcome the problems of GP, such as \emph{bloat} \citep{silva2008controlling} (the individual's complexity explosion over the generations) or convergence \citep{10.5555/1595536.1595590} issues, for example, with multi-objective strategies \citep{Tamaki1996Multi} or a partial derivative-based error fitness \citep{schmidt2009distilling}.

Driven by the current need for more interpretable models, non-GP-based methods have been developed to tackle SR. These methods propose to take advantage of the computational capacity of neural networks \citep{hornik1989multilayer} while providing an interpretable solution. For instance, they offer to encode the expression in the neural network structure and activation functions \citep{sahoo2018learning, kim2020Integration}, to predict a string expression \citep{anjum2019novel} with Recurrent Neural Networks, or to use Deep Reinforcement Learning (Deep RL) as a search engine \citep{petersen2021deep}. By combining partial derivative and neural networks, AI Feynman method \citep{udrescu2020ai} propose to make use of simplifying properties (such as units, symmetry, separability.. etc.) intrinsically present in physical expressions to repeatedly cut the global SR problem into simpler ones with fewer variables. Other methods that do not use neural networks rely on bayesian optimization \citep{jin2019bayesian} or perform nonlinear basis function expansion \citep{McConaghy2011}. 

However, as designed, these methods do not insert custom knowledge and expertise into the symbolic expression construction. 

\subsection{Knowledge Insertion by Constraints}
Because the search space in SR is very large and can leads to local optima, it is relevant to restrict the function space by removing sub-optimal search regions.
In line with Koza's work on GP, a preliminary solution called Strongly Typed GP (STGP) \citep{montana1995strongly} proposed to enforce data types constraints for computer program search, which decreases the search time and improves the generalizability of the found solutions. Regarding SR for physical laws (re)discovery, other kinds of constraints can be considered, such as physical units. More precisely, as each variable comes with its physical unit, arbitrarily combined variables can produce illegal unit combinations. To this end, dimensionally awareness GP \citep{keijzer1999dimensionally} was initially proposed to take into account unit knowledge in GP by minimizing the distance to a legal unit in the fitness function. 
 Note that, more than just constraining the search space, these constraints enforce structured knowledge and expertise about the problem within the learning. This knowledge can also take the form of ontologies \citep{prieschl2019using} to include prior knowledge as additional input features. However, both dimensionally-aware GP \citep{keijzer1999dimensionally} and ontology-guided GP \citep{prieschl2019using} do not ensure to produce dimensionally valid expressions. 

Toward this goal, the definition of explicit constraints can guarantee to produce only legal expressions with respect to the constraints. These constraints can be grammatical \citep{whigham1995grammatically} or ontological \citep{a14030076}. In Grammar-Guided Genetic Programming (G3P) \citep{whigham1995grammatically}, also called Grammar-Based GP, a Context-Free Grammar (CFG) \citep{CREMERS197586} is used to define constraint rules. Grammatical rules allow defining physical units, thanks to which G3P has found a variety of industrial applications \citep{10.1007/978-3-030-67667-4_27, cherrier2019consistent}. CFGs are often written in Backus-Naur form (BNF) \citep{DBLP:journals/cacm/Knuth64a}, which is made of: 
 \begin{itemize}
     \item \emph{Non-Terminals} also called \emph{symbols} (ex. $<$s$>$). One of the non-terminals is called \emph{start symbol}.
    \item \emph{Terminals}, character strings to replace non-terminals (ex. ``a", ``b", ``+"). The set of terminals can contain input features and operators to combine features.
    
    \item \emph{Rules} which defines how the terminals and non-terminals are connected. A set of rules for a specific symbol is called a \emph{production rule} where rules are separated by a vertical line $\rvert$, and $::=$ means ``defined as" 
    (for example $<$s$>$ ::= ``a" $\rvert$ ``b" $\rvert$ $<$s$>$+$<$s$>$).

\end{itemize}
Eventually, a grammar contains multiple production rules, one per symbol. To select which rule will replace a given symbol $<$symbol$>$, uniform sampling is made in $<$symbol$>$ production. McKay et al. \citep{mckay2010grammar} provides a detailed survey of G3P strategies and genetic operations. Because of its ability to restrict the symbol space, CFG structures have been a privileged topic of study. Outside SR, CFGs found other applications, such as in Grammar Variational Autoencoder \citep{kusner2017grammar} where they are used for symbolic data representation along with molecule representation. 

Nowadays, probabilistic CFG (noted PCFG) \citep{Sakakibara2017Probabilistic} are prefered as they allow to weigh the importance of a rule. In addition to CFG rules, they assign a probability to each rule in a production rule so that all probabilities for a given symbol add up to 1.
At the end of a production rule, a list preceded by a keyword \texttt{probs} and a double vertical line $||$ defines the probabilities associated to the rules.  The production rule structure now becomes: 
\begin{Verbatim}[fontsize=\small]
<symbol> ::= rule1 | rule2 |...|| probs [prob_r1,prob_r2,...]
\end{Verbatim}
PCFGs are of particular importance as they allow estimating the probabilities associated with each rule and discard unuseful rules. Probability distributions can be updated according to sampled expressions using Linear Genetic Programming
\citep{10.1145/3071178.3071325} or Monte Carlo sampling \citep{Brence2021Probabilistic}. 

\subsection{Reinforcement Learning}
Reinforcement Learning (RL) is a Machine Learning approach to solving Markov Decision Processes (MDPs), where the MDP is mainly defined by its \emph{state} space, \emph{action} space, \emph{state transitions probabilities}, and \emph{reward}.
In the RL paradigm \citep{sutton2018reinforcement}, an agent learns to achieve a task by interacting with its environment at discrete time steps. At each time step, the agent chooses an action among available actions according to a given policy. Next, the action is sent to the environment, which gives back a  reward feedback and a new state to the agent. Then, the agent can learn from the received reward signals to improve its policy. RL strategies are mainly valued-based like Deep Q-Network \citep{mnih2015humanlevel}, or policy-based like REINFORCE algorithm \citep{williams1992simple}.  Value-based RL learns a value function and deduce a policy from values. In contrast, policy-based RL explicitly learns a policy $\pi$ and keeps it in memory during learning. In Deep Reinforcement Learning (Deep RL), value and policy functions are approximated using neural networks. For example, the REINFORCE algorithm \citep{williams1992simple} at its core uses the policy gradient theorem to update the probability distribution of actions.
Actor-Critic algorithms \citep{konda2000actor} are inheriting from both strategies by trying to learn alongside a policy and its value to reduce variance and improve converge. 

RL has been applied to a variety of domains in pattern recognition \citep{pinol2012feature, khurana2018feature, 8476633}, from feature construction \citep{khurana2018feature} to feature-based aggregation \citep{8476633}. Including knowledge in RL is of paticular importance especially for safety issues \citep{alshiekh2018safe} where shielding strategy is used to correct actions if the chosen one causes a violation of some specified sort. However, few works have focused on building a symbolic representation for data. Recent work on this topic includes the use of RL to search among a library of operators and features for SR \citep{petersen2021deep} or the creation of symbolic computer programs \citep{pmlrv80verma18a}. 

Regarding interpretability, Deep Learning and RL approaches mostly produce complex solutions that are often hard to interpret. It is especially true in environments where most agents perform according to a black-box policy. To make these black-box approaches more grey and learn more interpretable RL policies, recent work proposes to learn symbolic policies either with GP \citep{hein2018interpretable} or Deep RL \citep{landajuela2021discovering}. However, even if they have interpretable outputs, most of these solutions do not include domain-related knowledge within the learning to ensure that the output solutions follow the prior knowledge.

\section{Reinforcement Based Grammar Guided Symbolic Regression (RBG2-SR)}
\label{sec:proposed_approach}

\subsection{Definition of the Reinforcement Learning Environment}
In this work, we adopt a Markov Decision Process (MDP) \citep{sutton2018reinforcement} modeling of the SR problem. More precisely, we choose to consider a Partially-Observable Markov Decision Process (POMDP)\citep{KAELBLING199899} in a finite episodic setting with maximum horizon $H$. 
This section is dedicated to the definition of the main components of the MDP in a RL setting, namely: \emph{state} and \emph{action} spaces and \emph{reward}. In Section 
\ref{subsubsec:grammar_definition}, we define  State and action spaces for SR and Section \ref{subsubsec:reward_definition} details the reward definition along with its properties. Section \ref{subsubsec:pomdp} combine the definitions from Sections 
\ref{subsubsec:grammar_definition} and \ref{subsubsec:reward_definition} to define the whole POMDP. Finally, Section \ref{subsubsec:policy_optimisation} details how to learn a policy over action, used to generate the action probabilities given the current state. 

 \subsubsection{Grammatical state and action space}
\label{subsubsec:grammar_definition}
\begin{figure}[ht]
    \centering
    \includegraphics[width=\linewidth]{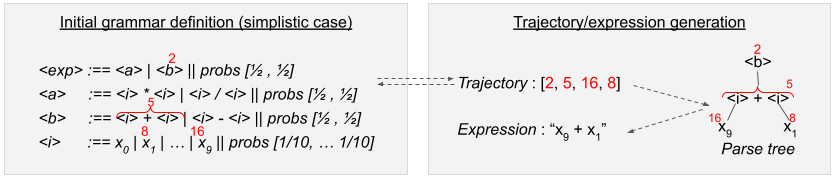}
    \caption{``$x_9 + x_1$" Expression and trajectory generation from a given grammar. A simplistic grammar is given (left), with actions numbered from 1 to 16 and start symbol $<exp>$. On the right, 
    a \emph{trajectory} is sampled from this grammar, from which we define the corresponding \emph{parse tree} and symbolic \emph{expression}}
    \label{fig:tree_generation_from_grammar}
\end{figure}

Let us consider a RL setting where the overall task is to find an optimal symbolic function $f^*$ so that $f^* = \operatorname*{argmin}_{f \in F_G} \lvert\lvert y - f(X) \rvert\rvert $, with $F_G$ being the function space accessible from a given grammar $G$. The grammar is defined by the tuple $(\sigma_{start}, (\sigma_{nt})_{nt \in NT}, (\sigma_t)_{t \in T}, \rho, \Psi)$ with $\sigma_{start}$ the start symbol of the grammar, $(\sigma_{nt})_{nt \in NT}$ a set of non-terminals, $(\sigma_t)_{t \in T}$ a set of terminals, $\rho$ the production rules to combine terminals/non terminals, and $\Psi$ the probability associated to each production rule. Figures \ref{fig:tree_generation_from_grammar} and \ref{fig:simplistic_grammar_example} in Section \ref{sec:experiments} show example BNF Grammar inspired by the work of Sotto and Melo \citep{10.1145/3071178.3071325}. 

Given this notation, we propose to define the construction of $f^*$ as a sequential
decision making problem where an agent sequentially chooses rules in the grammar to build up the function $f$.
In this grammatical space, we first specify the maximal number of steps to create $f$, called the maximal horizon $H$. We then define for each step $h \in [0, ..., H]$ an \emph{action}  $a_h$ as the \emph{selection of a rule} in the production rule accessible from the current state. We also propose to define the \emph{state} $s_h$ at step $h$ by $s_h=(a^{past}_h, a^{parent}_h, a^{siblings}_h, d_h, \sigma_h, m_h, \eta_h) $, with $a^{past}_h=[a_0, ..., a_{h-1}]$ all previously selected action, $a^{parent}_h$ the action taken by the parent in the parse tree, $a^{siblings}_h$ the action taken by each already computed siblings in the parse tree, $d_h$ the depth of the expression tree at step $h$, $\sigma_h$ the current type of symbol to find at step $h$, $m_h$ a mask over accessible actions from the current state symbol $\sigma_h$ and $\eta_h$ hidden information about the current state. We call sibling nodes those nodes with the same depth as the current node, and parent node the node above the current node in the parse tree. 
Alternative state definitions will be tested in Section \ref{subsubsec:ablation_study}. Initially, we define the state by the start symbol $\sigma_{start}$, previously selected actions is an empty list, $m_0$ masks out inaccessible actions from the initial symbol, and hidden information $\eta_0$ is randomly initialized.
A \emph{trajectory} of actions $\tau_k=(a_0^k, ..., a_h^k), h \in [H]$ is associated to each symbolic function $f_k$. We consider a case where the grammar is chosen and constructed so that there is always at least one action accessible at each step.

\begin{algorithm}[ht]
\caption{Single episode sampling, returns one function $f$ per episode. }\label{alg:episode_sampling}
\begin{algorithmic}[1]
\Require{maximal horizon $H$, policy $\pi_\theta$, grammar $G$}
\Function{Sample Episode}{$H$, $\pi_\theta$, $G$}
\State $queue, a^{past}, a^{parents}, a^{siblings}, f \gets$ empty
\State $d \gets$ 0
\State $\sigma \gets \mathtt{get\_start\_symbol}(G)$
\State $m \gets \mathtt{get\_mask}(\sigma, G)$
\State $\eta \gets \mathtt{random\_initialisation}()$
\For{h in H}
 \State $state \gets ((a^{past}, a^{parent}, a^{siblings}, d, \sigma, m, \eta) $ 
 \State $action\_probs, \eta \gets \pi_\theta(state) $ \Comment{$\pi_\theta$ is described in Figure \ref{fig:nn_architecture}(a)}
 \State $action \gets \mathtt{sample}(action\_probs) $\Comment{Corresponds to "Action Sampling" blue box in Figure \ref{fig:nn_architecture}}
 \State $a_{past} \gets \mathtt{append}(a_{past}, action)$
 \State $\sigma_{NT}^{child}, \sigma_{T}^{child} \gets \mathtt{get\_child\_symbols}(action, G)$
 \State $f \gets \mathtt{translate}(f, \sigma_{NT}^{child}, \sigma_{T}^{child})$
 \State $queue \gets \mathtt{extend}(\sigma_{NT}^{child}, queue)$ \Comment{Put $\sigma_{NT}$ at the start of the queue}
 \State $\sigma \gets \mathtt{pop}(queue)$
 \State $a^{parent}, a^{siblings} \gets \mathtt{get\_parent\_and\_siblings}(\sigma$, $a\_{past}$)
 \State $m \gets \mathtt{get\_mask}(G, \sigma$)
 \State $d \gets d + 1$
\EndFor

\Return $f$
\EndFunction

\end{algorithmic}
\end{algorithm}

The Algorithm \ref{alg:episode_sampling} details the episode sampling procedure. 
A visual example of this expression sampling is provided in Figure \ref{fig:tree_generation_from_grammar}. From a given grammar with 16 actions and $<$exp$>$ as start symbol, we generate a trajectory of actions. The trajectory construction goes as follow: 
\begin{enumerate}
    \item We begin by selecting the first action among accessible actions from the start symbol $<exp>$: either actions 1 or 2.  Given the probabilities associated with these actions, we perform a weighted sampling with the probabilities listed in \texttt{probs} as weights. Action 2 is sampled with the rule ``$<b>$". As this rule contains the non-terminal symbol $<b>$ we need to replace it with a rule from the grammar ``accessible" for the symbol $<b>$.
    \item The $3^{rd}$ row in the grammar defines actions accessible from $<b>$: actions 5 or 6. Given the weights, we sample action 5, ``$<i> + <i>$". It contains two non-terminal symbols ($<i>$ and $<i>$) that need to be replaced in the next steps. 
    \item The non-terminal symbol replacement is performed on one symbol at a time in a depth-first search manner, by looping over the first symbol until reaching a terminal symbol. We iterate this procedure until either the maximal trajectory length is reached or all non-terminal symbols encountered in all action selections have been replaced by a terminal value.
\end{enumerate}

\subsubsection{Reward definition}
\label{subsubsec:reward_definition}
The standard metric to minimize in SR is the Mean Squared Error (MSE). To match the RL definitions where the reward function is a monotonically increasing function, we use the squashing function $\frac{1}{1 + x}$.

\begin{equation}
\label{eq:reward}
    r_h  = \left\{
            \begin{array}{ll}
                 0 & if\ h < H \\
                 R = \frac{1}{1 + MSE(y, \hat{y})} & if\ h = H
            \end{array}
        \right. 
\end{equation}

  As the function $f$ can only be evaluated at the end of the episode when the function is complete, the reward $r$ equals 0 until the final step is reached and values  $\frac{1}{1 + MSE(y, \hat{y}) }$ at step $H$, where $\hat{y} = f(X)$ as shown in Equation \ref{eq:reward}. 
We also note the expected cumulative reward $R$ and highlight that $R=\sum_h{r_h}=r_H$. The sparsity property is used in Section  \ref{subsubsec:cost_function} to simplify the REINFORCE algorithm \citep{williams1992simple} loss function.

\subsubsection{Partially-Observable Markov Decision Process}
\label{subsubsec:pomdp}
 Given the previous space, action and reward definitions, the POMDP we consider here is defined by a tuple $(S, A, r, P, H, \Omega, O)$ with $S$ the state space, $A$ the action space, $r : S \times A \xrightarrow{} [0, 1]$ the reward function, $P : S \times A \xrightarrow{} [0,1]$ the transition kernel, $\Omega = (o_1, o_2, ..., o_K) $ a set of observations and $O$ a set of conditional observation probabilities $O(o\rvert s',a)$. We write as $P(s'\rvert s, a)$ the probability of having a transition to state $s \in S$ when taking action $a$ in state $s$. A POMDP setting is here considered to mitigate the fact that action effects are uncertain until the final state is reached and that the state might be partially observable. Each episode $k$ builds up a trajectory of actions  $\tau_k=(a_k^h)_{h \in [H]}$ that ends either when the maximum horizon is reached or when the function $f_k$ constructed by $\tau_k$ is complete and can be evaluated.

\subsubsection{Policy optimization}
\label{subsubsec:policy_optimisation}
Given the grammatical action space defined in Section \ref{subsubsec:grammar_definition}, there exists an optimal policy $\pi^*$ capable of generating a trajectory as close as possible to the symbolic function $f^*$. Depending on the choice of a judiciously constructed grammar, it is even possible to generate an optimal trajectory exactly corresponding to $f^*$. 

To search for this optimal symbolic function, we propose to learn a policy $\pi_\theta$, parametrized by a vector $\theta$ to generate the weights of grammatical action rules at each step $h\in [H]$ of the trajectory, from which we sample the next action. More precisely, to build $f$, the stochastic policy $\pi_{\theta}$ assigns a probability vector to accessible actions from a given state. At each step $h$, the action is then sampled according to the probability vector given by $\pi_{\theta}(s_h, o_h)$. Using the reward defined in Section \ref{subsubsec:reward_definition}, we iteratively update the parameters of $\pi_{\theta}$ to sample, in the future, more relevant trajectories with respect to the defined reward.

\begin{figure}[ht]
    \centering
    \includegraphics[width=\linewidth]{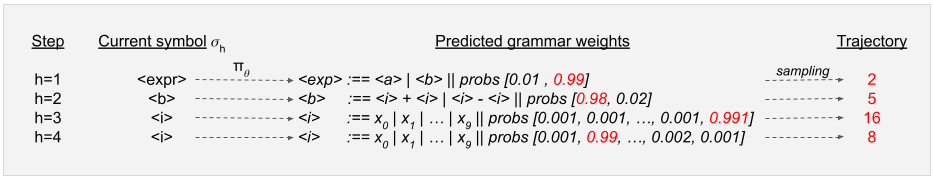}
    \caption{Weights and trajectory generation with the grammar from Figure \ref{fig:tree_generation_from_grammar} after training. The red elements define the results of the sampling on the grammar. Grammar weights are updated by searching for the target expression ``$x_9 + x_1$"}
    \label{fig:trained_grammar}
\end{figure}

Using the same grammar example from Figure \ref{fig:tree_generation_from_grammar}, we show in Figure \ref{fig:trained_grammar} how this grammar could be updated when trained on the search of the expression ``$x_9 + x_1$". After training, all unnecessary grammatical rules now have a low or zero probability, and it is easier to generate the correct target expression with this grammar.
 
 \subsection{Learning $\pi_{\theta}$ and finding $f^*$: Implementation Details}

\subsubsection{POMDP Modeling with a Recurrent Neural Network}
\begin{figure}
    \centering
    \includegraphics[width=0.7\linewidth]{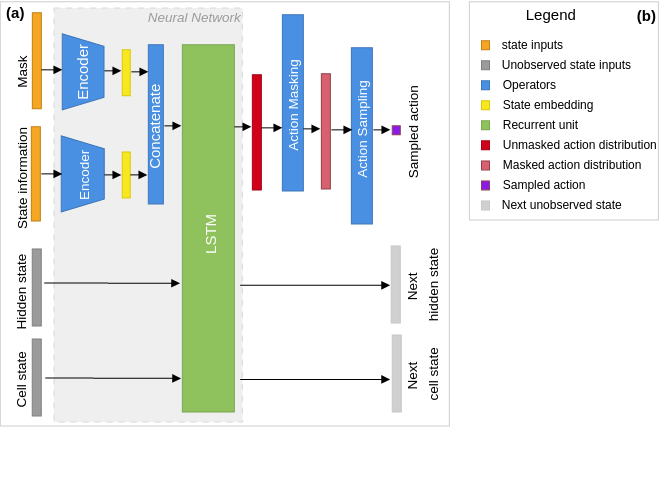}
    \includegraphics[interpolate=false,width=0.6\linewidth]{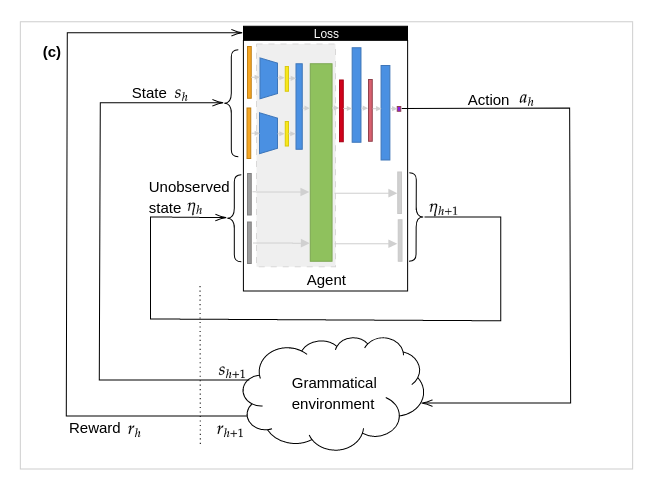}
    \caption{Neural Network architecture to learn $\pi_{\theta}$ (Better seen in color). \textbf{(a)} shows the recurrent architecture to predict an action from a given state as described in Algorithm \ref{alg:episode_sampling} lines 9 and 10. \textbf{(b)} is the color legend we use in \textbf{(a)} and \textbf{(c)}. \textbf{(c)} represents the POMDP agent-environment interaction}
    \label{fig:nn_architecture}
\end{figure}

 In order to find an optimal function $f^*$, we propose to use $\pi_{\theta}$ as an exploration tool, trained to emphasize exploration on the most relevant regions of the grammatical space. To do so, we learn $\pi_{\theta}$ with the 
 policy-gradient (PG) algorithm called REINFORCE \citep{williams1992simple} on the neural network architecture described in Figure \ref{fig:nn_architecture}. In Figure \ref{fig:nn_architecture}(a), we describe (using the legend in Figure \ref{fig:nn_architecture}(b)) the recurrent architecture to predict an action $a_h$ from a given state $s_h$ at step $h$. To handle recurrency we choose Long Short-Term Memory (LSTM) \citep{hochreiter1997long} networks. They are a purposely designed structure to capture long term temporal dependencies and they are also able to model the partial observability of the above-mentioned MDP \citep{10.1007/978-3-540-74690-4_71}.

 The neural network takes as input the state $s_h$ (in orange), and hidden state $\eta_h$ (in grey). All state-input features are encoded either using convolutions or feed-forward layers depending on their shape. The encoded state segments are then concatenated and handed to a LSTM cell along with the hidden state $\eta_h$. This recurrent cell both outputs an estimation of the observations for the next hidden state $\eta_{h+1}$, and an encoding of all actions in the grammar. This actions encoding is then masked using the state mask $m_h$ (see Section 
  \ref{subsubsec:invalid_action_masking}) and outputs the distribution over actions \emph{accessible} from the current state. Then, we sample action $s_h$ according to the distribution over accessible action.
 
\subsubsection{Invalid action Masking}
  \label{subsubsec:invalid_action_masking}
  To handle grammatical constraints on the action space, we propose to use at each step $h$ a mask over inaccessible actions $m_h$ in the current state $s_h$. Similarly to what is done by  \cite{huang2020closer}, invalid actions are masked through element-wise multiplication with a large negative value and followed by \texttt{softmax} function.

\subsubsection{Exploration Driven Cost Function}
 \label{subsubsec:cost_function}
REINFORCE algorithm \citep{williams1992simple} objective function, simplified with Equation 
\ref{eq:reward}, is 
$ J_{\theta} = \mathbb{E}_{\pi}$\textbf{\large [}$R \sum_h\ log\ \pi_{\theta}(a_h\rvert s_h)$\textbf{\large ]} from which we specify the optimal policy $\pi_{\theta}^* = \text{argmax}_{\theta} J_{\theta}$. 
As PG methods such as REINFORCE tends to have a large variance \citep{chung21a}, it is common practice to subtract a baseline to the batch, such as a moving average of rewards across batches. However, SR is only interested in building a policy that maximizes the best-performing trajectories found during training, and these strategies might have a slow convergence because many sampled trajectories are irrelevant and have a low or zero final reward. Based on these comments,  \cite{petersen2021deep} proposed a \emph{risk-seeking policy gradient}, which only compute the cost function based on the top$-\epsilon$ quantile of the expected rewards $R_{\epsilon}$, i.e., the most relevant trajectories of the batch: 
\begin{equation}
     J^{risk}_{\theta}(\epsilon) = \mathbb{E}_{\tau \sim \pi_{\theta}}\textbf{\Large [}(R(\tau\rvert \theta) -R_{\epsilon}) \sum_{h=0}^H \ log\ \pi_{\theta}(a_h\rvert s_h) \big\bracevert R(\tau\rvert \theta) > R_{\epsilon}\textbf{\Large ]}
\end{equation}
They also added a entropy term $\mathcal{H}$, weighted by $\lambda_\mathcal{H}$, to encourage exploration:
\begin{equation}
     J^{entropy}_{\theta}(\epsilon) = \mathbb{E}_{\tau \sim \pi_{\theta}}\textbf{\Large [} \mathcal{H}(\tau \rvert \theta)\ \big\bracevert R(\tau\rvert \theta) > R_{\epsilon}\textbf{\Large ]}
\end{equation}

Eventually, our final cost function becomes:
\begin{equation}
\begin{array}{rcl}
        J_\theta^{tot}(\epsilon)  &=& J^{risk}_{\theta}(\epsilon) +\lambda_\mathcal{H} J^{entropy}_{\theta}(\epsilon)\\ 
        &=&\mathbb{E}_{\tau \sim \pi_{\theta}}\left[
        \begin{array}{l}
            (R(\tau\rvert \theta) -R_{\epsilon}) \sum_{h=0}^H \ log\ \pi_{\theta}(a_h \rvert s_h) \\
            + \lambda_\mathcal{H} \mathcal{H}(\tau \rvert \theta)
        \end{array} \bigg\bracevert R(\tau\rvert \theta) > R_{\epsilon}\right]

\end{array}
\end{equation}

Given this cost function, the Neural network architecture is trained using the Algorithm \ref{alg:training}.
\begin{algorithm}[ht]

\caption{Training procedure }\label{alg:training}
\begin{algorithmic}[1]
\Require{$X$, target $y$, maximal horizon $H$, batch size $B$, number of training iterations $N$, quantile threshold $\epsilon$}
\State $\pi_\theta \gets \pi_{\theta_0}$
\For{n in N}
 \State $episodes \gets \mathtt{sample\_episodes}(H,\pi_{\theta}, G, B)$
 \Comment{Use Algorithm \ref{alg:episode_sampling} B times}
 \State $rewards \gets \mathtt{evaluate\_episodes}(episodes, X, y)$
 \State $best\_episodes\_and\_rewards \gets \mathtt{filter\_episodes}(episodes, rewards, \epsilon)$
 \State $\pi_\theta \gets \mathtt{update\_policy}(\pi_\theta, best\_episodes\_and\_rewards)$
\EndFor
\end{algorithmic}
\end{algorithm}

\subsubsection{Brief summary of RBG2-SR method}
To summarize, our proposed RBG2-SR method performs constrained SR where a grammatical structure restricts symbolic expressions creation. We adopt a POMDP setting with a finite horizon $H$, for which the associated reward $r_h$ is zero until the full expression $f$ has been generated.  A grammar defines a set of constraint rules used for the expression construction, which masks the non-accessible actions at each step $h \in [H]$. The weights of the actions accessible at the current time step are generated by $\pi_\theta$, a neural network learned with the REINFORCE algorithm, using a risk-seeking with entropy term loss.

\section{Experiments}
\label{sec:experiments}
In this section, we describe two experiments. The first one compares the proposed method with several state-of-the-art \citep{whigham1995grammatically,10.1145/3071178.3071325} SR solutions on reference benchmarks \citep{uy2011semantically, keijzer2003improving, 4632147, pagie1997evolutionary}. In the second experiment, we present a feature exploration use case where we show an application of our approach on a real-world dataset with an unknown relationship to uncover. Both data and code for this benchmark are freely available on Github \footnote{\href{https://github.com/laure-crochepierre/reinforcement-based-grammar-guided-symbolic-regression}{https://github.com/laure-crochepierre/reinforcement-based-grammar-guided-symbolic-regression}}.

\subsection{Experiment 1: Benchmark evaluation}
\subsubsection{Benchmarked methods and datasets}

To evaluate our method and compare it to other state-of-the-art works, we consider 34 functions gathered from the Nguyen \citep{uy2011semantically} (noted N1 to N10), Keijzer \citep{keijzer2003improving} (K1-15), Vladislavleva \citep{4632147} (V1-8), and Pagie \citep{pagie1997evolutionary}
benchmark suites with varying levels of difficulty.  The Nguyen benchmark suite is known to be the easiest one because it is mainly using one input feature and does not require optimizing for constant values in the expressions. Keijzer and Vladislavleva benchmarks are more complex as they contain functions with up to 5 inputs variables and also require to represent scaling constants. We also used Pagie (P1) \citep{pagie1997evolutionary} function, which has the reputation of being more challenging \citep{10.1145/2330163.2330273}. We applied several guidelines identified for Symbolic Regression benchmarking as provided by \cite{10.1145/2330163.2330273}. Our data generation procedure uses their functions and sampling intervals for train and test sets \citep{10.1145/2330163.2330273}.

The selected symbolic functions are benchmarked against the following grammar based methods: 
\begin{description}
    
    \item[Grammar Guided Genetic Programming](G3P) \citep{whigham1995grammatically} is a grammar guided genetic algorithm build using the Deap library. \citep{DEAP_JMLR2012}
    \item[Probabilistic Model Building Genetic Programming](GB-LGP) \citep{10.1145/3071178.3071325} updates the probability distribution  of a grammar according to selected individual from an evolutionary population in gradient-descent like algorithm.
\end{description}

As these methods are not directly available with our expression representation, we have re-implemented both methods. The following code is available on our Github repository.

\begin{figure}[ht]
    \centering
    \begin{Verbatim}[fontsize=\footnotesize]
<e> ::= (<e><dop><e>) | (<etw1><dopw1><etw1>) | <sop>(<e>) | <et> || probs [1/4,1/4,1/4,1/4]
<et> ::= (- x[x.columns<varidx>]]) | x[<varidx>] || probs [0.5, 0.5]
<etw1> ::= ( - x[<varidx>]) |  x[<varidx>] | 1 || probs [1/3,1/3,1/3]
<dopw1> ::=  + |  - || probs [1/2, 1/2]
<dop> ::= + | - | * | / || probs [1/4, 1/4, 1/4, 1/4]
<sop> ::= cos | sin | exp | log || probs [0.25, 0.25, 0.25, 0.25]
<varidx> ::= 1... nvar || probs [1/nvar ... 1/nvar]
    \end{Verbatim}
    \caption{Grammar example inspired by \citep{10.1145/3071178.3071325}. $<$\texttt{e}$>$ is the start symbol, $T = \{<$\texttt{e}$>,<$\texttt{et}$>, <$\texttt{etw1}$>,..., <$\texttt{varidx}$>\}$ NT=$\{$\texttt{x[], +,-,*,/,cos,sin, exp, log,1,..,nvar}$\}$}
    \label{fig:simplistic_grammar_example}
\end{figure}

To have comparable results between these methods, we propose to use the grammar described in Figure \ref{fig:simplistic_grammar_example} for G3P, GB-LGP, and RBG2-SR (ours). It describes the transitions between 7 symbols and defines the action space to search into, made of at least 19+$n_{var}$ actions (with $n_{var}$ the number of features in the dataset). Datasets are generated using the drawing process detailed in Appendix \ref{appendix:generation}.

After hyperparameters search using a grid search method for all three algorithms, we choose the following best perfoming parameters for RBG2-SR: $\lambda_H = 0.005$ and a learning rate $\alpha = 0.001$. All methods are compared on a maximal horizon of 50 actions and each run is performed on a population/batch of $1000$ expressions with a total of $2$ millions expressions tested at most (corresponding to a $2000$ iterations: batch size $\times$ nb training steps$ = 1000 \times 2000 = 2M$). 

\subsubsection{Benchmark results}
Expressions found by these methods on 30 independent runs are compared using Mean Squared Error (MSE) averages and standard deviations between the exact expression to uncover and best in-run solution of each algorithm.  The last row of the table corresponds to the result of the Mann–Whitney U test for independent samples \citep{Mann1947OnAT}. U-test results are summarised by counting the number of times each method performs better (symbol $+$), equivalently ($\sim$), or worse ($-$) than other methods.  ``Equivalently" refers to the case where two (or more) methods are equally good, and have the same best results. A method is said to perform ``worse" if at least one of the two other methods is performing ``better" or ``equivalently". Results for these benchmarks are shown in Table \ref{tab:exp1_benchmark_results} with best results per function in bold and equivalent scores in italic.
\begin{table}[ht]
\tiny
\setlength{\tabcolsep}{5pt}
\begin{center}
\begin{minipage}{\textwidth}
\caption{Mean Squared Error and Standard Deviation scores for benchmarked methods, averaged over 30 runs (best results in bold). The symbol $-$ is used when unable to compute a solution or when the solution error is larger than $10^{10}$. On the last two rows, is shown first the average MSE across all valid runs (out of 30 runs), and all benchmarks and then the count of times where each method is performing better (symbol $+$), equivalently ($\sim$) or worse ($-$) than others performed using the Mann–Whitney U test}
\label{tab:exp1_benchmark_results}
\begin{tabular*}{\textwidth}{@{}lccc@{}}
    
    \toprule
         Name & GB-LGP \citep{10.1145/3071178.3071325} & G3P \citep{whigham1995grammatically} & RBG2-SR (Ours)\\
         \midrule
N1&$5.71\times10^{-2}\ (\pm7.6\times10^{-2})$&$2.35\times10^{-3}\ (\pm2.6\times10^{-3})$&$\boldsymbol{0.00\ (\pm0.0)}$\\ 
N2&$9.29\times10^{-2}\ (\pm1.7\times10^{-1})$&$2.19\times10^{-2}\ (\pm5.8\times10^{-2})$&$\boldsymbol{0.00\ (\pm0.0)}$\\ 
N3&$2.24\times10^{-1}\ (\pm2.7\times10^{-1})$&$1.82\times10^{-2}\ (\pm2.4\times10^{-2})$&$\boldsymbol{0.00\ (\pm0.0)}$\\ 
N4&$2.24\times10^{-1}\ (\pm4.1\times10^{-1})$&$\boldsymbol{\mathit{1.48\times10^{-2}\ (\pm1.6\times10^{-2})}}$&$\boldsymbol{\mathit{1.62\times10^{-2}\ (\pm1.6\times10^{-2})}}$\\ 
N5&$6.16\times10^{-3}\ (\pm1.2\times10^{-2})$&$1.31\times10^{-3}\ (\pm1.9\times10^{-3})$&$\boldsymbol{5.87\times10^{-4}\ (\pm8.6\times10^{-4})}$\\ 
N6&$1.92\times10^{-2}\ (\pm1.4\times10^{-2})$&$1.70\times10^{-3}\ (\pm1.4\times10^{-3})$&$\boldsymbol{2.78\times10^{-4}\ (\pm7.9\times10^{-4})}$\\ 
N7&$1.48\times10^{-2}\ (\pm1.5\times10^{-2})$&$\boldsymbol{\mathit{3.97\times10^{-4}\ (\pm3.0\times10^{-4})}}$&$\boldsymbol{\mathit{3.30\times10^{-4}\ (\pm3.1\times10^{-4})}}$\\ 
N8&$2.11\times10^{-2}\ (\pm4.1\times10^{-2})$&$5.64\times10^{-2}\ (\pm2.9\times10^{-1})$&$\boldsymbol{1.24\times10^{-4}\ (\pm6.6\times10^{-4})}$\\ 
N9&$1.41\times10^{-1}\ (\pm1.2\times10^{-1})$&$\boldsymbol{\mathit{3.93\times10^{-2}\ (\pm1.1\times10^{-1})}}$&$\boldsymbol{\mathit{1.66\times10^{-2}\ (\pm2.3\times10^{-2})}}$\\ 
N10&$1.81\times10^{-2}\ (\pm3.1\times10^{-2})$&$6.67\times10^{-3}\ (\pm1.2\times10^{-2})$&$\boldsymbol{1.38\times10^{-3}\ (\pm1.5\times10^{-3})}$\\ 
K1&$3.38\times10^{-2}\ (\pm5.2\times10^{-3})$&$1.11\times10^{-2}\ (\pm9.7\times10^{-3})$&$\boldsymbol{2.35\times10^{-3}\ (\pm3.0\times10^{-3})}$\\ 
K2&$4.48\times10^{-2}\ (\pm1.1\times10^{-3})$&$3.72\times10^{-2}\ (\pm4.5\times10^{-3})$&$\boldsymbol{2.28\times10^{-2}\ (\pm7.4\times10^{-3})}$\\ 
K3&$4.50\times10^{-2}\ (\pm1.1\times10^{-4})$&$4.03\times10^{-2}\ (\pm6.5\times10^{-3})$&$\boldsymbol{3.15\times10^{-2}\ (\pm7.7\times10^{-3})}$\\ 
K4&$8.71\times10^{-2}\ (\pm2.0\times10^{-2})$&$2.44\times10^{-2}\ (\pm1.7\times10^{-2})$&$\boldsymbol{1.19\times10^{-2}\ (\pm8.9\times10^{-3})}$\\ 
K5&$-$&$-$&$-$\\ 
K6&$2.56\times10^{-2}\ (\pm7.8\times10^{-2})$&$\boldsymbol{1.46\times10^{-3}\ (\pm2.0\times10^{-3})}$&$2.32\times10^{-3}\ (\pm1.9\times10^{-3})$\\ 
K7&$\boldsymbol{\mathit{4.70\times10^{-4}\ (\pm1.8\times10^{-3})}}$&$6.59\times10^{-7}\ (\pm3.0\times10^{-6})$&$\boldsymbol{\mathit{0.00\ (\pm0.0)}}$\\ 
K8&$5.05\times10^{-1}\ (\pm1.0)$&$1.94\times10^{-1}\ (\pm2.1\times10^{-1})$&$\boldsymbol{2.92\times10^{-2}\ (\pm8.9\times10^{-2})}$\\ 
K9&$1.03\times10^{-3}\ (\pm3.8\times10^{-3})$&$\boldsymbol{3.11\times10^{-6}\ (\pm4.3\times10^{-6})}$&$4.08\times10^{-6}\ (\pm3.4\times10^{-6})$\\ 
K10&$2.45\times10^{-3}\ (\pm2.3\times10^{-3})$&$6.94\times10^{-4}\ (\pm7.5\times10^{-4})$&$\boldsymbol{1.79\times10^{-4}\ (\pm2.0\times10^{-4})}$\\ 
K11&$8.49\times10^{-1}\ (\pm1.9)$&$\boldsymbol{5.52\times10^{-1}\ (\pm2.8\times10^{-1})}$&$2.42\ (\pm9.5)$\\ 
K12&$3.40\times10^{+2}\ (\pm4.4\times10^{+2})$&$6.33\times10^{+4}\ (\pm3.5\times10^{+5})$&$\boldsymbol{2.36\ (\pm1.2)}$\\ 
K13&$ - $&$3.04\ (\pm3.6)$&$\boldsymbol{5.23\times10^{-1}\ (\pm1.2)}$\\ 
K14&$5.65\times10^{-1}\ (\pm7.5\times10^{-2})$&$4.21\times10^{-1}\ (\pm1.9\times10^{-1})$&$\boldsymbol{1.49\times10^{-1}\ (\pm1.7\times10^{-1})}$\\ 
K15&$2.41\ (\pm1.5)$&$2.03\times10^{+8}\ (\pm1.1\times10^{+9})$&$\boldsymbol{9.22\times10^{-1}\ (\pm1.6\times10^{-1})}$\\ 
P1&$2.14\times10^{-1}\ (\pm2.5\times10^{-1})$&$\boldsymbol{\mathit{1.66\times10^{-1}\ (\pm1.2\times10^{-1})}}$&$\boldsymbol{\mathit{1.11\times10^{-1}\ (\pm9.2\times10^{-2})}}$\\ 
V1&$\boldsymbol{\mathit{5.74\times10^{-2}\ (\pm2.7\times10^{-2})}}$&$ - $&$\boldsymbol{\mathit{5.13\times10^{-2}\ (\pm1.7\times10^{-2})}}$\\ 
V2&$\boldsymbol{8.39\times10^{-2}\ (\pm1.9\times10^{-2})}$&$ - $&$1.47\times10^{-1}\ (\pm6.4\times10^{-1})$\\ 
V3&$ - $&$4.52\ (\pm1.7\times10^{+1})$&$\boldsymbol{8.31\times10^{-1}\ (\pm2.9\times10^{-1})}$\\ 
V4&$\boldsymbol{\mathit{3.83\times10^{-2}\ (\pm3.6\times10^{-3})}}$&$\boldsymbol{\mathit{3.87\times10^{-2}\ (\pm4.7\times10^{-3})}}$&$\boldsymbol{\mathit{3.71\times10^{-2}\ (\pm5.3\times10^{-3})}}$\\ 
V5&$2.77\times10^{-1}\ (\pm1.2\times10^{-1})$&$1.54\times10^{-1}\ (\pm9.1\times10^{-2})$&$\boldsymbol{4.10\times10^{-2}\ (\pm3.3\times10^{-2})}$\\ 
V6&$4.76\ (\pm5.3)$&$ - $&$\boldsymbol{8.64\times10^{-1}\ (\pm7.6\times10^{-1})}$\\ 
V7&$2.60\times10^{+1}\ (\pm7.6\times10^{+1})$&$1.13\times10^{+1}\ (\pm3.8)$&$\boldsymbol{9.94\ (\pm1.0)}$\\ 
V8&$4.12\ (\pm2.9\times10^{-1})$&$3.93\ (\pm1.9)$&$\boldsymbol{2.06\ (\pm5.7\times10^{-1})}$\\ 
\hline 
 Avg& \multirow{2}{*}{$1.14\times10^{+1}$}& \multirow{2}{*}{$1.70$}& \multirow{2}{*}{$ \boldsymbol{6.09\times10^{-1}}$}\\
 MSE & & & \\
\hline
U test & $+1\ /\sim3\ /-29$& $ +3/\sim5/-25$& $ \boldsymbol{+22 /\sim 7 / - 4}$\\
\botrule
\end{tabular*}

\end{minipage}
\end{center}
\end{table}

As shown on the antepenultimate row in Table \ref{tab:exp1_benchmark_results}, our method lower the average MSE by one order of magnitude and performs statistically better than other methods on 22 out of 33 benchmarks according to the Mann–Whitney U test. The K5 benchmark is not taken into account in testing because all methods perform poorly and produce a high MSE error on this benchmark (above $ 10^{14}$). We also note that we perform similarly to other methods for seven benchmarks, leading to an \emph{at least} similar error in more than $90\%$ of the tested benchmarks (30 out of 33 benchmarks).

More precisely, looking at the Nguyen benchmark, we show that our proposed method outperforms other methods for all functions, except for N4, N6, and N9, where the error is similar between RBG2-SR and G3P. Then, for Keijzer and Vladislavleva benchmarks, our method proposes solutions with a lower (resp. equivalent) error for 10 out of 15 (resp. 1/15) and 5 out of 8 (resp. 3/8) benchmarks. We also highlight that our method seems to complement the evolutionary G3P method, especially on the Keijzer benchmark, as G3P has a lower error on the few benchmarks where our method is weaker.

\begin{figure}[t]
    \centering
    \includegraphics[width=\linewidth]{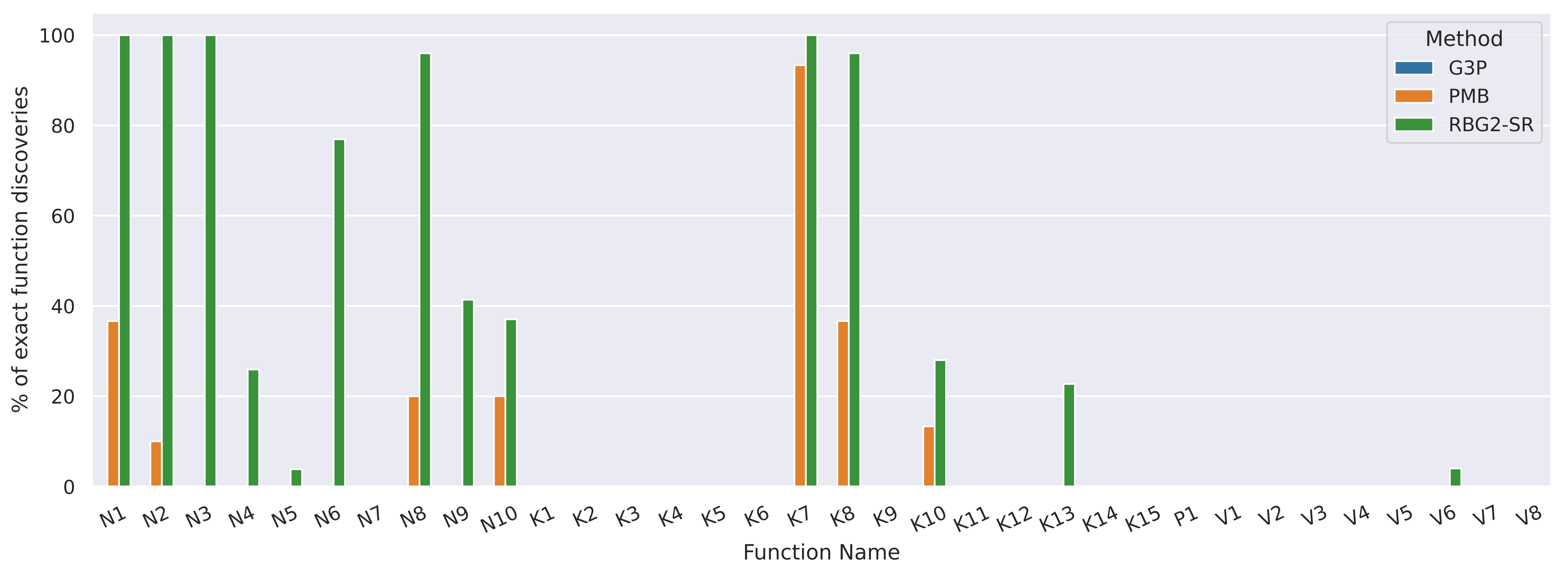}
    \caption{Percentage of exact solutions found on the four benchmarks for the three methods RBG2-SR (blue), GB-LGP (orange), G3P (green). Better seen in color. $100\%$ means that all $30$ runs uncover the solution, $50\%$ that $15$ out of the $30$ were able to match the right results}
    \label{fig:exact_percent_all}
\end{figure}
Let us now analyse the ability of our algorithm to precisely retrieve the exact target symbolic expression. It is also worth noting that our method reaches a zero error for several benchmarks (N1-3 and K7), meaning that we recover all the time the exact solution. More details on this result is described in Figure \ref{fig:exact_percent_all}. In this Figure, we represent for each benchmark and method the number of times (in percentage) where we can recover the exact solution up to the numerical precision.
First, we see that two of the three methods recover functions mostly on the Nguyen and Keijzer benchmarks. As we detailed above, unlike other methods for most benchmarks, the RBG2-SR method can often recover the exact solution for $9$ out of $10$ expressions of the Nguyen benchmark. On the same benchmark, G3P only recovers approximate solutions, and GB-LGP can only find four functions with a lower percentage. 
While GB-LGP recovers 3 out of 15 expressions from the Keijzer benchmark, our RBG2-SR method recovers 4 functions with a higher recovery rate. RGB2-SR also recover sometimes one function on the Vladislavleva benchmark.
Note that these results are tied to the grammar used. The GB-LGP and G3P methods could potentially find more exact solutions by choosing another grammar.  Moreover, among other unfound functions, some are difficult or even impossible to construct with the chosen grammar. This is especially true for functions such as K1-3 or V1 that require finding decimal constants, which is not supported in this version of the grammar.  

\subsubsection{Ablation study}
\label{subsubsec:ablation_study}
In order to identify which elements of the RBG2-SR method are essential to the success of the expression search, we performed an ablation study on elements of the state definition $s_h$ and on the algorithm. In the algorithm itself, we try removing the risk-seeking objective $J_\theta^{risk}$ (keeping all trajectories) and the entropy loss term $J_\theta^{entropy}$. Regarding the state definition, we compare state defined with and without: the current symbol $\sigma_h$, the current mask $m_h$, the current depth $d_h$, the parent node $a_h^{parent}$, the siblings nodes $a_h^{siblings}$ and the previously selected actions $a^{past}_h$. This ablation study uses the ten functions of the Nguyen benchmark and was run 10 times for each ablation/function combination. All results are compared to the \emph{baseline} case where no element is occluded. 
The results are summarized in Table \ref{tab:ablation_study}.

\begin{table}[ht]
\begin{center}
\begin{minipage}{0.85\textwidth}
    \caption{Ablation Study. Averaged MSE scores ($\pm$ Standard Deviation) and percentage of variation over 10 runs on the Nguyen benchmark. All results are to be compared to the baseline case: $Variation (\%) = 100 \frac{baseline - ablation}{baseline}$}
    \label{tab:ablation_study}
    \begin{tabular}{@{}lccc@{}}
    \midrule%
    Type & Ablation & MSE &Variation(\%)\\ 
    \hline
     &Baseline& $1.58\times10^{-3}$\ ($\pm4.92\times10^{-3}$) & -\\
    \hline\multirow{2}{*}{Algorithm}& No entropy & $1.51\times10^{-2}$\ ($\pm4.41\times10^{-2}$) & $860\% $\\ 
 &No risk-seeking& $5.16\times10^{-2}$\ ($\pm8.68\times10^{-2}$) & $3200\% $\\ 
    \hline
    \multirow{5}{*}{State} & No parent & $1.51\times10^{-3}$\ ($\pm4.68\times10^{-3}$) & $-4\% $\\
 & No siblings & $2.57\times10^{-3}$\ ($\pm8.75\times10^{-3}$) & $63\% $\\ 
 & No past actions & $2.97\times10^{-3}$\ ($\pm1.01\times10^{-2}$) & $87\% $\\ 
 & No depth & $3.73\times10^{-3}$\ ($\pm1.18\times10^{-2}$) & $140\% $\\ 
 & No symbol & $3.83\times10^{-3}$\ ($\pm1.23\times10^{-2}$) & $140\% $ \\
 \botrule
    \end{tabular}
    \end{minipage}
\end{center}
\end{table}

First, when looking at the algorithm learning itself, we show in Table \ref{tab:ablation_study} that the combination of the two loss terms is relevant to our problem. The MSE significantly increases when removing one of these terms. The risk-seeking policy is crucial to this type of learning: removing the risk-seeking policy increases the error by $3200\%$ when compared to the proposed method with risk-seeking. 
The entropy term is also of great importance, with an error increase of $860\%$.

We performed a second type of ablation on the state definition. In this part of the ablation study, we tried removing elements from \emph{state} inputs from the Neural Network (the orange block called ``state information"): either parent action information $a^{parent}$, siblings action information $a^{siblings}$, previously selected actions $a^{past}$. Results from Table \ref{tab:ablation_study} indicate that these siblings, past actions, depth, and symbols are highly beneficial to the expression search since removing one of these terms increases the error by at least 63\%. Depth and symbol information seems to be of almost equal importance for a successful expression search. 
Regarding parent information, it seems that removing this information tends to reduce the error by a small 4\%. However, by looking more carefully at each benchmark, removing the parent information is only beneficial to the search for benchmark N9. Except for N1-3 and N7 (where both configurations always recover the exact expression), the complete proposed algorithm (\emph{Baseline}) outperforms the parent ablation. For the remaining benchmarks, the corresponding increase lies between $+50$ and $+55\%$. With these results, we choose to keep parent information in the state definition. 

\subsection{Experiment 2: Interpretability analysis of a use case}
This work aims at describing an algorithm that provides interpretable symbolic solutions directly readable by humans. From the first experiment, we also see a potential application of our approach to more complex datasets with an unknown relationship between a set of observations $X$ and a target variable $y$, where $X$ and $y$ variables may be of different physical units. In this scenario, the use of a grammar is particularly important to restrict outputs to realistic solutions in terms of physical units. For example it is not physically feasible to add a speed (measured in meters per second) with a distance (in meters).

Toward this physical interpretability goal, we designed a second experiment on the Airfoil Self-Noise dataset \citep{brooks1989airfoil}, to compare the solutions proposed by our algorithm with the ones given by four state-of-the-art \citep{whigham1995grammatically,10.1145/3071178.3071325, McConaghy2011, petersen2021deep} methods. The dataset is accessible on the UCI Machine Learning Repository\footnote{\url{https://archive.ics.uci.edu/ml/datasets/airfoil+self-noise} (Accessed \today) }. The tested methods are GB-LGP, G3P from the previous experiment, and two other non-grammar based methods: 

\begin{description}
    \item[Fast Function Extraction](FFX) \citep{McConaghy2011} which applies pathwise learning to a large set of nonlinear functions, and exploits the path structure to generate models that trade-off error/complexity. 
    \item[Deep Symbolic Regression](DSR from \citep{petersen2021deep}) a deep learning algorithm which uses a RL approach to search the solution space without grammatical constraints. 
\end{description}

\subsubsection{Dataset Description}
The problem we focus on in this experiment is predicting scaled sound pressure level (SSPL) on different size NACA 0012 airfoils. The estimation is made using the following variables: frequency (unit $Hz$), angle of attack (degree $^{\circ}$), chord length (meters $m$), free-stream velocity (meters per second $m.s^{-1}$), suction side displacement thickness (meters $m$).The measurements are obtained using airfoils taken at various wind tunnel speeds and angles of attack. The span of the airfoil and the observer position is fixed for all measurements. The dataset is split between 70\% in the train set and 30\% in the test set.

 \subsubsection{Grammar construction}
The first preprocessing step, before methods comparison, is the definition of a constrained grammar that contains premice of knowledge on the studied topic, such as the one used in Figure \ref{fig:grammar_exp2}. 
The start symbol is $<$exp$>$. This symbol describes the dimensions (units) and structures the algorithm allows as a return. From the previous studies \citep{Lau2009ANN} on this data, we want to find an expression in the form: 
$\mathtt{exp=constant - 10*log10(child\_expression)}$.
We also add several other structures to leave to the algorithm the freedom to explore. 

\begin{figure}[ht]
    \centering
    \begin{Verbatim}[fontsize=\footnotesize]
<exp>      ::=  <unit> * const  | <no_unit> * const 
           | const-10*log10(<no_unit>/const)*<no_unit> 
           | const-10*log10(<unit>/const)*<no_unit>
           ||probs [0.25, 0.25, 0.25, 0.25]
<unit>     ::= <distance> | <velocity> | <time> |(<no_unit> * <unit>) 
           || probs [0.25,0.25,0.25,0.25]
<no_unit>  ::= <no_unit>*<no_unit>| cos(x.alpha) | sin(x.alpha) | <distance>/<distance> 
            | <velocity>/<velocity> | <time>/<time> ||  probs [0.16,0.16,...,0.16]
<velocity> ::= (<velocity><dop><velocity>) | (<distance>/<time>) | x.U_infinity 
            || probs [0.33,0.33,0.33]
<distance> ::= (<distance><dop><distance>) | (<velocity>*<time>) | abs(<distance>)  
            | x.delta | x.c || probs [0.2,...,0.2]
<time>     ::=(<distance>/<velocity>)| (1/x.f) || probs [0.5,0.5]
<dop>      ::= - | + || probs [0.5,0.5]
    \end{Verbatim}
    \caption{Grammar used in the second experiment on the Airfoil dataset. The start symbol is $<$exp$>$ }
    \label{fig:grammar_exp2}
\end{figure}
To sum up the grammar from Figure \ref{fig:grammar_exp2}, the three first lines describe what the units and non-units (composition of units) of the problem are. In the four final lines, the grammar constrains the operations on each dimension (or unit) to only physically consistent combinations. These lines also define how to go from one unit to another by using physical properties (such as velocity law). This part of the grammatical description is of particular importance to describe the expertise and knowledge we want to include to constrain the search space during the SR resolution. 

\subsubsection{Experiments and results}

\begin{table}[ht]
\setlength{\tabcolsep}{2pt}
\tiny
\begin{center}
\begin{minipage}{\textwidth}
\caption{Analysis of the Airfoil Self-Noise benchmark. Best expression found are presented along with their MSE, determination coefficient $\mathcal{R}^2$, and complexity $\mathcal{C}$ scores }
     \label{tab:exp2_results}
\begin{tabular*}{\textwidth}{@{}lccccc@{}}
        \toprule
        Method & Expression & MSE & $\mathcal{R}^2$& $\mathcal{C}$ &$\mathcal{C}-H\ (<0)$\\
        \midrule
        DSR & $\displaystyle
        - \alpha + U_{infinity} - \frac{U_{infinity}}{\operatorname{sin}(\operatorname{log_{10}}(U_{infinity}))}$ & $239.15$ & $-4.07$ & $10$&$-40\ (<<)$\\
        & & & & \\
        FFX & 
        \begin{tabular}{c}
        \shortstack{ $\displaystyle 0.001 U_{infinity}^{2} - 0.026 U_{infinity} + 14.4 \alpha \delta $\\ 
        $ - 0.492 \alpha + 12.8 c^{2} + ... + \operatorname{log_{10}}{\left(f \right)}  $\\ 
        $ + 18.8 \operatorname{log_{10}}{\left(\delta \right)} - 71.3 \operatorname{log_{10}}{\left(f \right)} + 272$ }\end{tabular} & $10.3$ & $\boldsymbol{0.78}$ & $140$&$90\ (>>)$\\
        & & & & \\
        G3P & $\displaystyle 101.38 - 10 \operatorname{log_{10}}{\left(\frac{1}{f} + \frac{\delta f^2 c^2  }{U_{infinity}^3} \right)}$ & $19.5$ & $0.58$ & $22$& $-28\ (<)$\\
        & & & & \\
        GB-LGP & $\displaystyle  127.36 -10 \operatorname{log_{10}}{\left(\frac{c \delta f}{\frac{U_{infinity} \delta}c  + 2 U_{infinity}} \right)} \cos^{2}{\left(\alpha \right)} $ & $35.6$ & $0.24$ & $24$& $-26\ (<)$ \\
        & & & & \\
        RBG2-SR (ours) &$\displaystyle \boldsymbol{83.85 - \frac{10 \operatorname{log_{10}}{\left(\frac{ U_{infinity}}{c f}\right) + \frac{U_{infinity}^2}{ f^2 c \delta}}}{1 + \frac{c \delta f^{2}}{U_{infinity}^2}}}$  & $13.0$& \emph{$0.72$} & $\boldsymbol{32}$&$\boldsymbol{-18\ (<)}$\\
        \botrule
    \end{tabular*}

\end{minipage}
\end{center}
\end{table}

In this experiment, all algorithms are run 10 times (except for FFX \citep{McConaghy2011} as it is deterministic), and the best expression of all runs is shown in Table \ref{tab:exp2_results}. Their results are compared on the test set based on the MSE error, determination coefficient $\mathcal{R}^2$ and complexity $\mathcal{C}$. The complexity $\mathcal{C}$ is defined by the sum of operations and input features used in the formula.
From Table \ref{tab:exp2_results}, we can note that the best performing algorithm on this dataset is FFX, with an error of $10.3$. However, this method also produces the expression with the highest complexity (around 140), above the maximal horizon $H$ of $50$. As seen in the column \emph{Expression}, FFX solution is complex, not directly readable by a human and combine variables with different units. 
Among other four methods, we highlight that our proposed RBG2-SR method produces the second lowest error and highest determination coefficient of $0.72$, while keeping an acceptable complexity of 36, close to the threshold $H$ but lower. Moreover, it is worth noticing that all grammar-based methods used a expression which uses the format :
$\mathtt{constant - 10*log10(child\_exp)}$. For example, the DSR method finds a simple solution, largely under the threshold $H$. However, this solution is too simplistic and doesn't respect dimensional consistency. These results tend to advocate for the usage of grammatical constraints for equation discovery. The expression found by our method, could for example be used to estimate the value of the $\mathtt{constant}$ in the above-mentioned equation. 

Eventually, regarding unit or dimensional consistency, all non-grammar based methods constructed forbidden combinations of different units, showing that their are not yet able to compete with grammar-based methods to build physically-relevant expressions.

\section{Conclusion and perspectives}
\label{sec:conclusion}


This study proposes a new algorithm (RBG2-SR) for Grammar Guided Symbolic Regression using a Reinforcement Learning search approach that allows the inclusion of domain knowledge within the learning process and in the format of the solutions. We describe a POMDP modeling of the SR task in a grammatical action space. The proposed method is benchmarked against grammar-based state-of-the-art algorithms and shows significant improvements over other algorithms regarding the error metric and exact expression discovery. We performed an ablation study of the blocks of our algorithm and state definition. The results show that parent, sibling, past actions, depth and symbol informations are all important elements of the state definition. 
In the second experiment, we also show how the use of a grammar based approach could be useful and interpretable when working on a dataset with physical constraint between input features. 

From the obtained results, we also foresee different perspectives to this work. First, when comparing G3P and RGB2-SR results, we envision improvements by doing cross-learning \citep{9533735} between G3P to encourage exploration and our method for learning and sampling. Moreover, as the grammar construction process can be a time-consuming task, we could draw inspiration from techniques that automatically build ontologies \citep{DBLP:journals/dke/EmaniSFG19} to automatically create and improve grammar.
We also foresee application perspectives for our method to find interpretable policies that follow expert grammatical rules. From the second experiment, we also want to explore the behavior of our algorithm when dealing either with longer horizons up to $100$ actions to create more expressive expressions or with shorter horizons for more concise and interpretable expressions.

\backmatter
\bmhead{Acknowledgments} This work was supported by the French Association Nationale de la Recherche et de la Technologie (ANRT) grant number 2018/1466. We also thank Benjamin Donnot for helpful discussions, Remy Cl\'ement and Baltazar Donon for their comments and suggestions.

\begin{appendices}
\section{Benchmark generation information}
\label{appendix:generation}

\setcounter{table}{3}
\renewcommand{\thetable}{\arabic{table}}
\renewcommand*{\theHtable}{\thetable}

SR benchmarks used in this paper: Nguyen \citep{uy2011semantically}, Keijzer \citep{keijzer2003improving}, Vladislavleva \citep{4632147}, and Pagie \citep{pagie1997evolutionary} (respectilvely noted N1-10, K1-15, V1-8 and P1).
Variables (column \emph{Vars}) are $x, y, z, v, w$ and their corresponding representation in the grammar is $x[1]$ to $x[5]$. $U[a,b,c]$ is a uniform sampling of $c$ samples between $a$ to $b$.  $E[a,b,c]$ samples in a grid of evenly spaced points with an interval of c, from a to b. Table  \ref{tab:nguyen_keijzer_vladislavleva_benchmarks} is  an extended version of the generation information presented by \cite{10.1145/2330163.2330273}.

\begin{table}[ht]
\tiny
\setlength{\tabcolsep}{5pt}

\begin{center}
\begin{minipage}{\textwidth}
\caption{Nguyen (respectilvely noted N1-10), Keijzer (respectilvely noted K1-15) Vladislavleva and Pagie (respectilvely noted V1-8 and P1) benchmarks}
\label{tab:nguyen_keijzer_vladislavleva_benchmarks}
\begin{tabular*}{\textwidth}{@{}lcccc@{}}
\toprule
         Name & Function & Vars & Train set & Test Set \\
         \hline
         N1 & $x^3+x^2+x$ & 1 & $U[0, 2, 20]$ & $U[0, 2, 20]$\\  
         N2 & $x^4+x^3+x^2+x$ & 1 & $U[-1, 1, 20]$ & $U[-1, 1, 20]$\\ 
         N3 & $x^5+x^4+x^3+x^2+x$ & 1 & $U[-1, 1, 20]$ & $U[-1, 1, 20]$\\ 
         N4 & $x^6+x^5+x^4+x^3+x^2 + x$ & 1 & $U[-1, 1, 20]$ & $U[-1, 1, 20]$\\ 
         N5 & $sin(x^2)cos(x)-1$ & 1 & $U[-1, 1, 20]$ & $U[-1, 1, 20]$\\ 
         N6 & $sin(x) + sin(x+x^2)$ & 1 & $U[-1, 1, 20]$ & $U[-1, 1, 20]$\\ 
         N7 & $ln(x+1) + ln(x^2+1)$ & 1 & $U[0, 2, 20]$ & $U[0, 2, 20]$\\ 
         N8 & $\sqrt(x)$ & 1 & $U[0, 4, 20]$ & $U[0, 4, 20]$\\ 
         N9 & $sin(x) + sin(y)$ & 2 & $U[0, 2, 100]$ & $U[0, 2, 100]$\\
         N10 & $2sin(x)cos(y)$ & 2 &$U[0, 2, 100]$ & $U[0, 2, 100]$\\ 
         K1 & $0.3xsin(2\pi x)$ & 1 & $E[-1, 1, 0.1]$ & $E[-1, 1, 0.001]$ \\
         K2 & $0.3xsin(2\pi x)$ & 1 & $E[-2, 2, 0.1]$ & $E[-2, 2, 0.001]$ \\
         K3 & $0.3xsin(2\pi x)$ & 1 & $E[-3, 3, 0.1]$ & $E[-3, 3, 0.001]$ \\
         K4 & $x^3e^{-x}cos(x)sin(x)(sin^2(x)cos(x)-1)$ & 1 & $E[0, 10, 0.05]$ & $E[0.05, 10.05, 0.05]$ \\
         \multirow{2}{*}{K5} & \multirow{2}{*}{$\frac{30xz}{(x-10)y^2}$} & \multirow{2}{*}{3} & $x,z:U[-1,1,1000]$ &  $x,z:U[-1,1,10000]$\\
         & & & $y:U[1,2,1000]$&  $y:U[1,2,10000]$\\
         K6 & $\sum_i^x \frac{1}{x}$ & 1 &$E[1, 50, 1]$ & $E[1, 120, 1]$ \\
         K7 & $ln(x)$ & 1 & $E[1, 100, 1]$ & $E[1, 100, 0.1]$\\
         K8 & $\sqrt(x)$ & 1 & $E[0, 100, 1]$ & $E[0, 100, 0.1]$\\
         K9 & $arcsinh(x)$ & 1 & $E[0, 100, 1]$ & $E[0, 100, 0.1]$\\
         K10 & $x^y$ & 2 & $U[0,1,100]$ & $E[0,1,0.01]$ \\
         K11 & $xy+sin((x-1)(y-1))$ & 2 & $U[-3, 3, 20]$ & $E[0,1, 0.01]$\\
         K12 & $x^4-x^3+\frac{y^2}{2}-y$ & 2 & $U[-3, 3, 20]$ & $E[0,1, 0.01]$\\
         K13 & $6sin(x)cos(y)$ & 2 & $U[-3, 3, 20]$ & $E[0,1, 0.01]$\\
         K14 & $\frac{8}{2+x^2+y^2}$ & 2 & $U[-3, 3, 20]$ & $E[0,1, 0.01]$\\
         K15 & $\frac{x^3}{5}+\frac{y^3}{2}-y-x$ & 2 & $U[-3, 3, 20]$ & $E[0,1, 0.01]$\\
         V1 & $\frac{e^{-(x-1)^2}}{1.2+(y-2.5)^2}$ & 2 & $U[0.3, 4, 100]$ & $E[-0.2, 4.2, 0.1]$ \\
         V2 & $e^{-x}x^3cos(x)sin(x)(sin^2(x)cos(x)-1)$ & 1 & $E[0.05, 10, 0.1]$ & $E[-0.5, 10.5, 0.05]$ \\
         \multirow{2}{*}{V3} & \multirow{2}{*}{$e^{-x}x^3cos(x)sin(x)(sin^2(x)cos(x)-1)(y-5)$} & \multirow{2}{*}{2} & $x:E[0.05,10,0.1]$ & $x:E[0.05,10,0.1]$\\
         & & & $y:E[0.05,10.05,2]$ & $y:E[-0.5,10.5,0.5]$\\
         V4 & $\frac{10}{5+(x-3)^2+(y-3)^2+(z-3)^2+(v-3)^2+(w-3)^2}$ & 5 & $U[0.05, 6.05, 1024]$ & $U[-0.25, 6.35, 5000]$\\
         \multirow{2}{*}{V5} & \multirow{2}{*}{$30\frac{(x-1)(z-1)}{y^2(x-10}$} & \multirow{2}{*}{3} & $x:U[0.05,2,300]$& $x:E[-0.05,2.1,0.15]$\\
         & & & $y:U[1,2,300]$ & $y:E[0.95,2.05,0.1]$\\
         V6 & $6sin(x)cos(y)$ & 2 & $U[0.1, 5.9, 30]$ & $E[-0.05, 6.05, 0.02]$\\
         V7 & $(x-3)(y-3) + 2sin((x-4)(y-4))$ & 2 & $U[0.05, 6.05, 300]$ & $U[-0.25, 6.35, 1000]$\\
         V8 & $\frac{(x-3)^4+(y-3)^3-(y-3)}{(y-2)^4+10}$ & 2 & $U[0.05, 6.05, 50]$ & $E[-0.25, 6.35, 0.2]$\\
         P1 & $\frac{1}{1+x^{-4}}+\frac{1}{1+y^{-4}}$ & 2 & $E[-5, 5, 0.4]$ & $E[-5, 5, 0.4]$ \\
         
\botrule
\end{tabular*}
\end{minipage}
\end{center}
\end{table}

\end{appendices}

\section*{Declarations}
\bmhead{Funding}
This work was supported by the French Association Nationale de la Recherche et de la Technologie (ANRT) [CIFRE convention between Universit\'e de Lorraine and Rte, grant number 2018/1466].
\bmhead{Conflict of interest/Competing interests} The authors declare that they have no conflict of interest.
\bmhead{Ethics approval} Not applicable
\bmhead{Consent to participate} Not applicable
\bmhead{Consent for publication} Not applicable
\bmhead{Availability of data and materials}
The experiment benchmarks are generated following the procedure detailed in Appendix \ref{appendix:generation} and data used in the second experiment can be found on the UCI Machine Learning Repository. Data generation code is available on our Github repository.
\bmhead{Code availability} The code is freely available on the Github repository whose link is shared in this document.
\bmhead{Authors' contributions} All authors contributed to the study conceptualization, design, and investigation. Data collection and analysis were performed by Laure Crochepierre. The first draft of the manuscript was written by Laure Crochepierre and all authors commented on previous versions of the manuscript. All authors read and approved the final manuscript.

\bibliography{biblio}

\end{document}